\documentclass[10pt,twocolumn,letterpaper]{article}
\PassOptionsToPackage{table}{xcolor}
\usepackage[pagenumbers]{cvpr}
\usepackage[margin=1in,columnsep=0.25in]{geometry}
\usepackage{times,xcolor}
\usepackage{graphicx,amsmath,amssymb,booktabs,placeins}
\usepackage{afterpage,stfloats}
\usepackage{tabularx,multirow}
\definecolor{cvprblue}{rgb}{0.21,0.49,0.74}
\definecolor{tableheader}{HTML}{E6E6E6}
\definecolor{tablehighlight}{HTML}{EAF4FF}
\usepackage{microtype}
\usepackage[pagebackref,breaklinks,colorlinks,allcolors=cvprblue]{hyperref}

\newcommand{\nrm}[1]{\left\lVert #1\right\rVert_2}
\newcommand{\unit}[1]{\overline{#1}}
\newcommand{\CE}{\operatorname{CE}}
\newcommand{\softplus}{\operatorname{softplus}}
\newcommand{\clip}{\operatorname{clip}}
\newcommand{\smean}{\operatorname{sceneMean}}

\usepackage{capt-of}

\usepackage{float}

\author{\normalsize Xinyang Zhang$^{1,*}$\quad Feiyu Han$^{2,*}$\quad Chenhao Wu$^{2,*}$\\[3pt]
\normalsize Yang Liu$^{3}$\quad Shang Luo$^{3}$\quad Yingguang Yang$^{3}$\\[3pt]
\normalsize Feng Zhang$^{4}$\quad Chongyang Zhang$^{5}$\quad Hao Zheng$^{5}$\\[3pt]
\normalsize Kefu Xu$^{3}$\quad Congjing Ran$^{6}$\quad Bin Chong$^{3,\dagger}$\\[3pt]
\small $^{1}$Huazhong University of Science and Technology\quad $^{2}$University of the Chinese Academy of Sciences\\[2pt]
\small $^{3}$Peking University\quad $^{4}$Fuzhou University\\[2pt]
\small $^{5}$Fullive-AI\quad $^{6}$Wuhan University\\[2pt]
\small $^{*}$Equal contribution.\quad $^{\dagger}$Corresponding author.\\[2pt]
\small Correspondence: \texttt{chongbin@pku.edu.cn}\\[2pt]
}

\title{OSLA: Observation-Aware Local Alignment and Cross-Scene Relation Learning for Any-Scenario Person Re-Identification}
\date{}
\begin{document}
\maketitle

\begin{abstract}
Any-Scenario Person Re-Identification (AS-ReID) aims to retrieve identities across spectral modalities and aerial–ground viewpoints using a unified model. However, changes in viewpoint and spectrum can make local correspondences unreliable, while conventional sampling may omit cross-camera positive relations needed for training. To address these challenges, we propose OSLA, an observation-aware framework built on supervised CLIP-ReID that combines Local Identity Alignment (LIA) and Cross-Scene Relation Learning (CSR). Specifically, LIA uses historical identity observations and competing-identity information to select reliable cross-scene local correspondences while allowing unmatched regions. Local-to-global distillation transfers identity knowledge from correctly classified historical local features into the global representation. CSR couples observed-cell sampling with spectrum-grouped ranking to jointly organize and optimize cross-camera relations. Sampling across two cameras and two spectra provides same- and cross-spectrum positives, while ranking constrains positive–negative scores separately by gallery spectrum on the final retrieval descriptor. Both modules operate only during training, preserving the original visual encoder and retrieval descriptor at inference. Experiments on WHU-MARS-1000 and WHU-MARS-2337 demonstrate that OSLA achieves absolute Rank-1 gains of 5.0\% and 4.8\% over UAD, respectively, under the AS-ReID protocol. The code will be available.
\end{abstract}

\begin{figure}[!t]
\centering\includegraphics[width=\columnwidth]{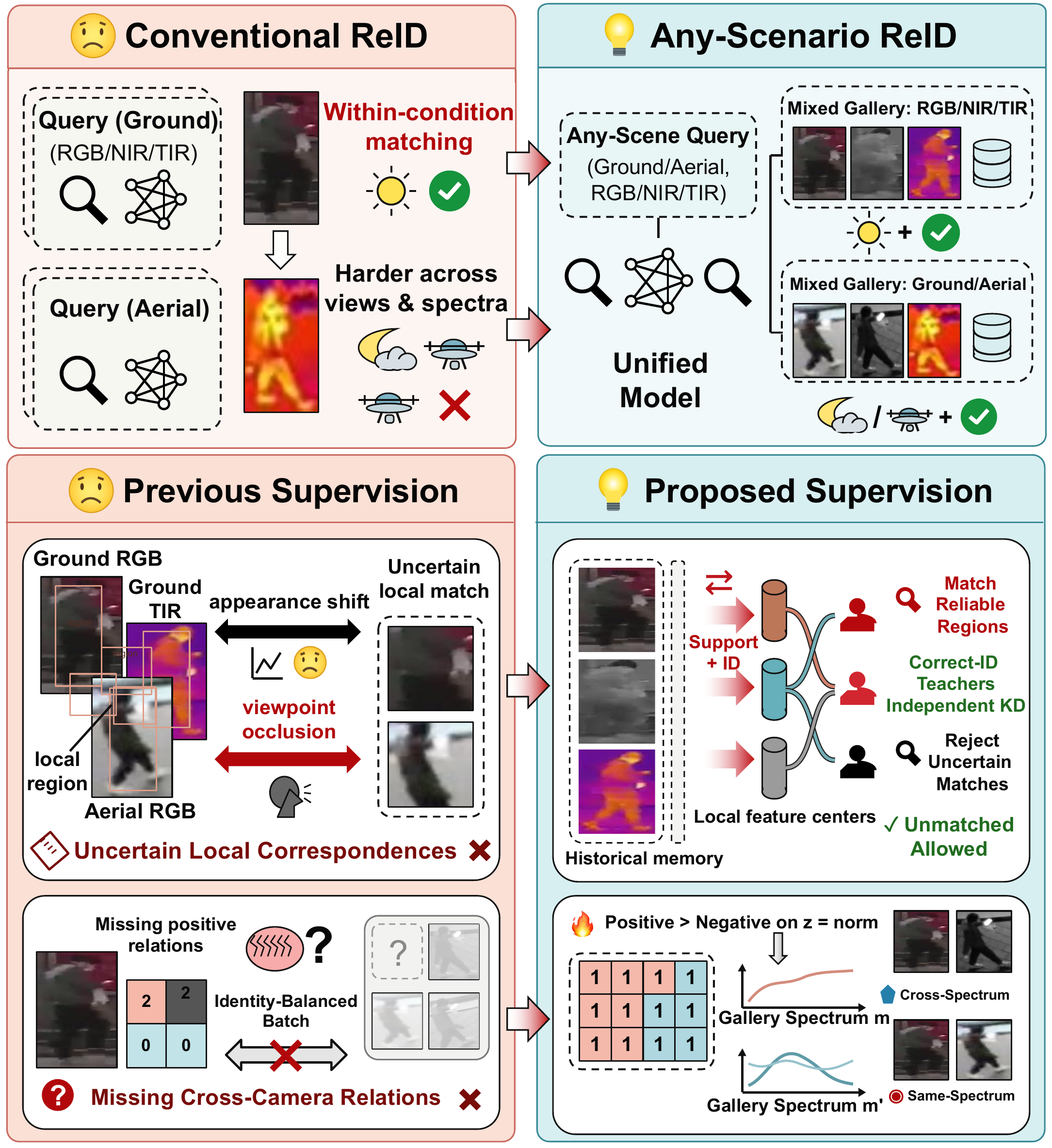}
\caption{Task setting, challenges, and training mechanisms. Top: conventional and any-scenario ReID. Bottom left: viewpoint and spectrum changes make local matches uncertain, while identity-balanced sampling may omit cross-camera positives. Bottom right: LIA permits unmatched regions and independently selects local teachers; CSR couples OCS sampling and spectrum-grouped ranking.}
\label{fig:taskoverview}
\end{figure}

\section{Introduction}
\label{sec:intro}
Person re-identification (ReID) aims to retrieve images of the same person across non-overlapping cameras~\cite{ye2022survey}. Any-scenario person re-identification (AS-ReID) extends this task to mixed observations from different sensing modalities and camera platforms~\cite{zhao2026whumars}. Given a query from any scene, a single model searches a gallery containing red--green--blue (RGB), near-infrared (NIR), and thermal-infrared (TIR) images captured by aerial and ground cameras. Aerial--ground viewpoint changes alter which body regions are visible and how they are arranged, while spectral differences change their appearance and discriminative value. Their interaction makes cross-scene identity matching more difficult than matching within a single acquisition condition.

Existing methods improve identity representations through shared feature learning~\cite{he2021transreid,zhang2024vdt}, local feature modeling~\cite{kim2023partmix}, and vision--language pretraining~\cite{radford2021clip,li2022clipreid}. However, learning a shared representation does not by itself specify which local observations should be aligned or which cross-camera relations should be included in training. We consider these two limitations in the AS-ReID setting.

The first difficulty concerns the reliability of local correspondence. Aerial--ground viewpoint changes can hide body regions, while spectral differences alter the appearance and informativeness of the visible regions. Similar local features may therefore arise from different identities, and the same identity may not provide a comparable region in another scene. As illustrated in the bottom-left panel of Fig.~\ref{fig:taskoverview}, forcing a match for every region can impose incorrect alignment targets. Historical observations offer additional references, but local similarity alone does not establish their suitability for supervision.

A second difficulty concerns the availability of cross-camera training relations. Identity-balanced sampling controls the number of images per identity, but does not explicitly ensure same- and cross-spectrum positives from another camera. The sampling example in Fig.~\ref{fig:taskoverview} illustrates this distinction: a batch can satisfy the identity-count constraint while missing the positive relations required for mixed-gallery retrieval. A ranking objective cannot constrain these missing relations. We therefore consider batch construction and ranking supervision together, rather than relying on identity balance alone.

In this paper, we propose Observation-Aware Local Alignment and Cross-Scene Relation Learning (OSLA), a unified training framework that combines selective local identity supervision with cross-scene relation learning. Local Identity Alignment (LIA) compares current local features with historical same-identity observations from other scenes. It screens references using observation support and identity competition while permitting unmatched assignments. Independently selected, correctly classified historical local features provide local-to-global distillation. Cross-Scene Relation Learning (CSR) complements this selective supervision with observed-cell sampling (OCS) and spectrum-grouped ranking (SGR). OCS supplies cross-camera same- and cross-spectrum positives, and SGR constrains positive--negative rankings separately by gallery spectrum on the final descriptor. Both modules operate only during training, using a single visual encoder and global descriptor at inference.

Our main contributions are summarized as follows:
\begin{itemize}
\item We introduce LIA, which selects historical cross-scene local correspondences using observation support and identity competition while allowing unmatched regions. Independently selected local teachers transfer identity knowledge to the global representation.
\item We introduce CSR, which couples observed-cell batch construction with spectrum-grouped ranking. It jointly supplies and optimizes cross-camera same- and cross-spectrum relations on the descriptor used at inference.
\item Experiments on WHU-MARS-1000 and WHU-MARS-2337 demonstrate the effectiveness of OSLA under AS-ReID, reaching 14.60\%/12.87\% mean average precision (mAP) and 34.47\%/30.49\% Rank-1, respectively. Component comparisons further examine the contributions of local alignment and relation learning.
\end{itemize}

\section{Related Work}
\label{sec:related}
\subsection{Multispectral Aerial--Ground ReID}
Aerial--ground ReID studies identity matching under substantial viewpoint changes. VDT~\cite{zhang2024vdt} decouples viewpoint-related features, DTST~\cite{wang2024dynamic} selects informative tokens, and SeCap~\cite{wang2025secap} adapts prompts for cross-view retrieval. Visible--infrared ReID instead focuses on spectral differences~\cite{wu2017rgbir,nguyen2017regdb}, using feature aggregation in DDAG~\cite{ye2020ddag}, pattern alignment in MPANet~\cite{wu2021mpanet}, and knowledge transfer in IDKL~\cite{ren2024idkl}. MP-ReID~\cite{ha2025multi} and AG-VPReID.VIR~\cite{nguyen2025agvpreidvir} further extend retrieval across platforms. WHU-MARS~\cite{zhao2026whumars} brings aerial--ground RGB/NIR/TIR observations into a unified AS-ReID setting, with UAD learning shared identity prototypes. Unlike approaches centered on global identity alignment, OSLA explicitly considers the reliability of local supervision and the availability of cross-camera relations within training batches.

\subsection{Local Features and Correspondence}
Local features preserve regional identity cues. PCB~\cite{sun2018pcb} and MGN~\cite{wang2018mgn} learn parts; HACNN~\cite{li2018hacnn} learns attention, while HOReID~\cite{wang2020horeid} and PAT~\cite{li2021pat} address partially visible persons. PartMix~\cite{kim2023partmix} regularizes cross-modal part learning, and ViSA~\cite{zhang2026visa} incorporates view-aware queries and local graphs into its retrieval representation. Following rejectable assignment in SuperGlue~\cite{sarlin2020superglue}, LIA permits unmatched regions. Knowledge distillation transfers teacher predictions~\cite{hinton2015distilling} or inter-example relations~\cite{park2019rkd}. LIA transfers predictions from independently selected historical local teachers to the global representation.

\subsection{Adaptation and Metric Learning}
Transfer learning can regularize parameters toward a pretrained model~\cite{li2018l2sp}. For vision--language models, CLIP~\cite{radford2021clip} supports adaptation through learned contexts in CoOp~\cite{zhou2022coop} and CoCoOp~\cite{zhou2022cocoop}, or visual prompts~\cite{jia2022vpt}. We adopt two-stage identity-context learning and visual optimization~\cite{li2022clipreid}, and combine them with observation-aware local supervision and cross-scene relation learning. For metric learning, triplet~\cite{schroff2015facenet,hermans2017triplet}, Multi-Similarity~\cite{wang2019multisimilarity}, Circle~\cite{sun2020circle}, and supervised contrastive~\cite{khosla2020supcon} objectives organize positive and negative relations; cross-batch memory~\cite{wang2020xbm} expands the available historical candidates. CSR instead couples observed camera--spectrum coverage with spectrum-grouped ranking on the final descriptor.

\begin{figure*}[!t]
\centering\includegraphics[width=\textwidth]{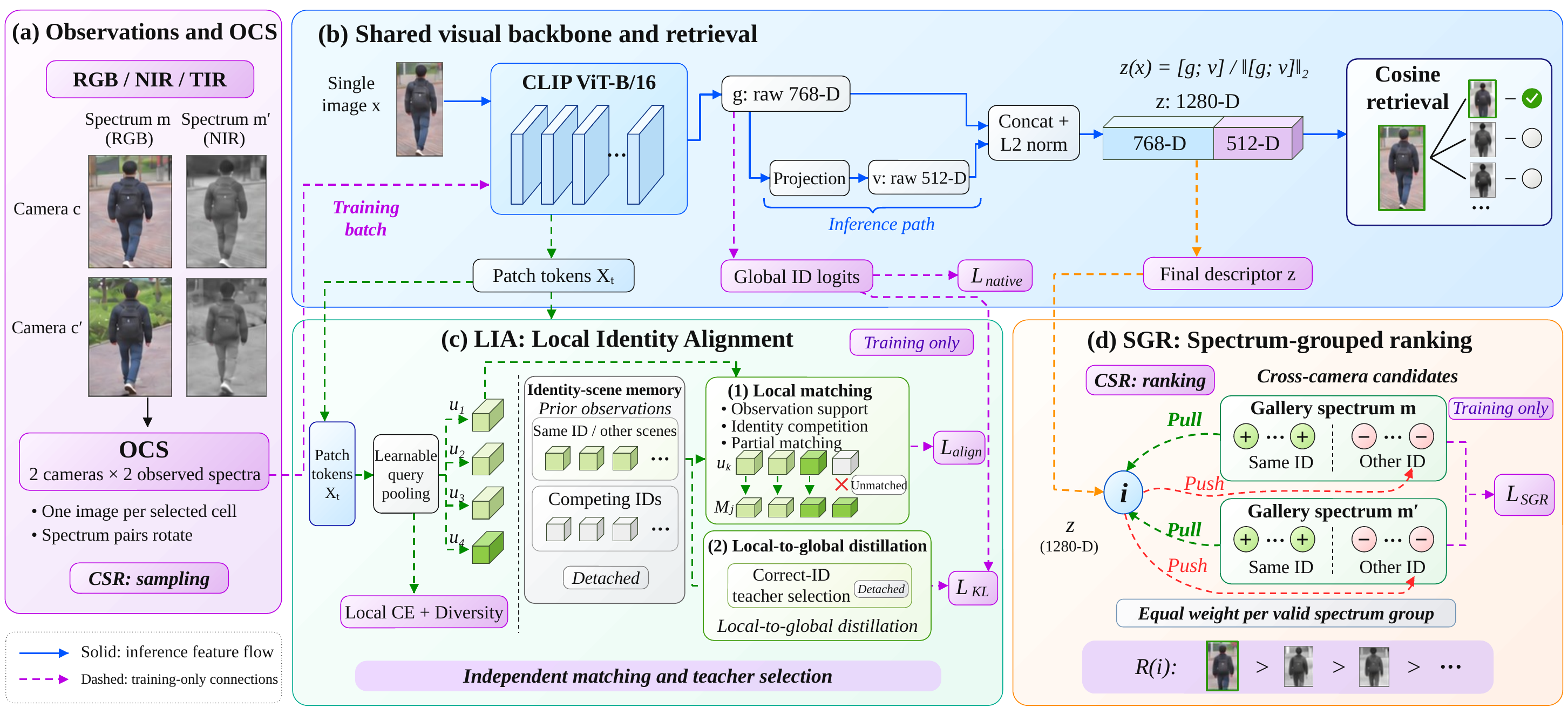}
\caption{OSLA architecture. Solid blue arrows show the inference feature flow: a single image yields a normalized 768+512-D descriptor. Dashed links are training-only. (a) OCS constructs batches; (b) the shared backbone extracts features; (c) LIA performs selective matching and independent local-to-global distillation; (d) SGR optimizes the final descriptor. OCS and SGR together form CSR.}
\label{fig:framework}
\end{figure*}

\section{Method}
\label{sec:method}
\subsection{Framework}
\label{sec:framework}
\paragraph{Problem setup.}
Given a query image $x_q$, AS-ReID uses a single model to retrieve the same identity from a mixed gallery covering all scenes. A scene is a platform--spectrum combination $s=(p,m)$, where $p\in\{\mathrm G,\mathrm A\}$ denotes ground or aerial platforms and $m\in\{\mathrm{RGB},\mathrm{NIR},\mathrm{TIR}\}$ denotes the imaging spectrum. The training set is $\mathcal D_{\rm tr}=\{(x_i,y_i,c_i,p_i,m_i)\}$, with identity label $y_i$, camera label $c_i$, and scene $s_i=(p_i,m_i)$. The observed camera--spectrum combinations for identity $y$ are
\begin{equation}
\Omega_y=\{(c_i,m_i):y_i=y\}.
\label{eq:observedcells}
\end{equation}
Scene labels encode platform and spectrum, whereas camera labels identify image sources. Different cameras may share a scene. We therefore organize local memory by identity and scene, and use camera labels additionally for sampling and ranking.

The goal is to learn a unified representation $z(x)$ that preserves identity across these acquisition conditions. Scene and camera labels organize training supervision; they are not inputs to the inference descriptor.

\paragraph{Overall architecture.}
The overall OSLA framework is shown in Fig.~\ref{fig:framework}. We combine LIA and CSR in a unified visual training framework to address uncertain local correspondences and the availability of cross-scene batch relations, respectively.

We adopt a two-stage training strategy~\cite{li2022clipreid}. Stage 1 freezes the CLIP visual and text encoders and learns text contexts for training identities. Stage 2 fixes the identity text features and optimizes the visual encoder through identity classification, triplet, and image--text objectives. For image $x$, the Vision Transformer (ViT)~\cite{dosovitskiy2021vit} variant ViT-B/16 takes $256\times128$ images with patch stride 16 and produces patch tokens $\{X_t\}$, a 768-dimensional classification ([CLS]) token feature $g$, and its 512-dimensional projection $v$. The retrieval descriptor concatenates the two global features and then normalizes them:
\begin{equation}
z(x)=\frac{[g;v]}{\nrm{[g;v]}}.
\label{eq:descriptor}
\end{equation}
LIA extracts local information from patch tokens and assists global identity learning through selective alignment and distillation. CSR combines OCS with SGR to directly optimize $z(x)$. Both modules act on the shared visual encoder only during training. Inference ranks gallery images by cosine similarity of $z(x)$ without local branches, memory, or a text encoder. Camera labels are used only for evaluation filtering, not descriptor extraction.

\subsection{Local Identity Alignment}
\label{sec:local}
Local regions provide identity cues, but their visibility and appearance vary across scenes. We introduce LIA to select historical local evidence for alignment and global identity learning. As shown in Fig.~\ref{fig:framework}(c), it permits unmatched regions and independently selects local teachers, so useful identity evidence need not depend on an accepted correspondence.

\paragraph{Local features and memory.}
We first aggregate patch tokens with $K$ learnable queries. Unlike fixed spatial partitions, these queries do not assign a predefined body part to each local feature. Queries $q_{ks}$ adapt to the scene through shared, spectrum, and platform components. Local features are
\begin{equation}
\alpha_{kt}=\operatorname{softmax}_t\!\left(\unit q_{ks}^{\top}\unit X_t/\tau_q\right),
\quad u_k=\sum_t\alpha_{kt}X_t,
\label{eq:query}
\end{equation}
where an overbar denotes $\ell_2$ normalization and $\tau_q$ is the attention temperature. Spectrum-specific classifiers provide local identity supervision, while an attention-overlap penalty encourages different queries to attend to different regions.

An identity--scene memory stores historical local centers $M_{ysk}$ and their observation support and dispersion. We assign each eligible reference a reliability weight $\rho_{ysk}$ that increases with observation support and decreases with dispersion. This favors stable cross-scene evidence over sparsely observed or variable features.

\paragraph{Competition and rejectable matching.}
Observation support describes the stability of a reference, but does not establish whether it distinguishes the identity from others. We therefore compare same-identity candidates with competing-identity references before computing a correspondence. Let $C_{kj}=\unit u_k^\top M_{ys'j}$, and let $H_{ks'}$ be the largest similarity between the current feature and eligible memory centers of other identities in scene $s'$. Competition is evaluated against other identities with eligible historical references. A candidate must satisfy
\begin{equation}
C_{kj}\geq\eta,\qquad C_{kj}-H_{ks'}>\epsilon_{\rm id},
\label{eq:eligibility}
\end{equation}
where $\eta$ is a similarity threshold and $\epsilon_{\rm id}$ requires an advantage over competing identities. The first condition requires sufficient same-identity similarity; the second requires that similarity to exceed the strongest available competing-identity reference.

For each source scene, we augment the score matrix with unmatched assignments and use optimal transport~\cite{sarlin2020superglue} to compute soft matches. Eligible similarities define candidate scores, while an unmatched option lets a region decline unsupported correspondence. Sinkhorn normalization~\cite{cuturi2013sinkhorn} and row/column capacity constraints yield weights $T_{is'kj}$. The alignment loss for image $i$ is
\begin{equation}
\ell_i^{\rm align}=
\frac{\sum_{s',k,j}\rho_{y_is'j}T_{is'kj}
(1-\unit u_{ik}^{\top}M_{y_is'j})}
{\sum_{s',j}\rho_{y_is'j}}.
\label{eq:align}
\end{equation}
The denominator includes all eligible source weights, so rejecting matches weakens supervision rather than amplifying a small accepted subset. $\mathcal L_{\rm align}$ averages eligible images within scenes and then across scenes. Historical references provide fixed targets for learning the current local features.

\paragraph{Local-to-global knowledge distillation.}
Local matching determines whether regions provide comparable evidence, whereas distillation transfers identity predictions. A historical region may recognize the identity correctly even when its counterpart is not visible in the current image. We therefore select teachers independently of the matching assignments. We classify eligible foreign-scene memory centers using their spectrum-specific classifiers and retain only centers whose identity prediction is correct. Let $\mathcal V_i$ denote the teacher centers. Their predictions are aggregated by source weight:
\begin{equation}
\pi_i=\frac{\sum_{(s',j)\in\mathcal V_i}\rho_{y_is'j}
\sigma(b_{y_is'j}/T_{\rm KD})}
{\sum_{(s',j)\in\mathcal V_i}\rho_{y_is'j}},
\label{eq:teacher}
\end{equation}
where $b_{y_is'j}$ are classification logits, $\sigma$ is softmax, and $T_{\rm KD}$ is the distillation temperature. The identity logits $a_i$ of the current 768-dimensional global feature are optimized through knowledge distillation~\cite{hinton2015distilling} using the Kullback--Leibler (KL) divergence:
\begin{equation}
\mathcal L_{\rm KL}=\smean_{i:\mathcal V_i\ne\varnothing}
T_{\rm KD}^2\operatorname{KL}\!\left(\pi_i\,\middle\|\,\sigma(a_i/T_{\rm KD})\right).
\label{eq:kl}
\end{equation}
Here $\smean$ first averages images with teachers within each scene and then averages valid scenes. The independent teacher criterion allows identity supervision from local features whose correspondence is rejected. Images without teachers contribute no distillation term.

\subsection{Cross-Scene Relation Learning}
\label{sec:relation}
LIA selects historical evidence for local supervision, but does not determine which cross-camera positives occur in the current batch. We introduce CSR to address this complementary issue by combining observed-cell sampling (OCS) with spectrum-grouped ranking (SGR). OCS supplies same- and cross-spectrum positives from different cameras. SGR then constrains their rankings against negatives within each gallery spectrum, directly supervising the descriptor used for retrieval.

\paragraph{Observed-cell sampling.}
For each identity, OCS selects two cameras and two spectra such that all four combinations exist:
\begin{equation}
\begin{gathered}
\{(c,m),(c,m'),(c',m),(c',m')\}\subseteq\Omega_y,\\
c\ne c',\qquad m\ne m'.
\end{gathered}
\label{eq:rectangle}
\end{equation}
One image is sampled per selected cell, supplying each anchor with same- and cross-spectrum positives from the other camera. Only observed cells are sampled, with four images per identity. This organizes cross-camera relations within the same four-image budget.

\paragraph{Spectrum-grouped ranking.}
A mixed gallery contains candidates with different spectral appearances. We group ranking constraints by gallery spectrum, comparing each positive with negatives from the same spectrum rather than combining all candidates into one group. For anchor $i$ and spectrum $m$, define
\begin{equation}
\begin{aligned}
\mathcal P_{im}&=\{j:y_j=y_i,\ c_j\ne c_i,\ m_j=m\},\\
\mathcal N_{im}&=\{j:y_j\ne y_i,\ c_j\ne c_i,\ m_j=m\}.
\end{aligned}
\label{eq:relations}
\end{equation}
Both sets exclude same-camera images, consistent with evaluation filtering. Let $z_i=z(x_i)$ and $s_{ij}=z_i^\top z_j$. For groups with nonempty positive and negative sets, log-mean-exp aggregates negative scores:
\begin{equation}
h_{im}=\tau_{\rm SGR}\!\left[
\log\sum_{n\in\mathcal N_{im}}e^{s_{in}/\tau_{\rm SGR}}
-\log|\mathcal N_{im}|\right].
\label{eq:negative}
\end{equation}
This emphasizes hard negatives while reducing the influence of their number. The within-group ranking loss is
\begin{equation}
\ell_{im}=\frac{1}{|\mathcal P_{im}|}\sum_{p\in\mathcal P_{im}}
\softplus\!\left(\frac{\mu_{\rm SGR}+h_{im}-s_{ip}}{\tau_{\rm SGR}}\right),
\label{eq:group}
\end{equation}
The margin $\mu_{\rm SGR}$ encourages positive scores to exceed the aggregated negative score, while $\tau_{\rm SGR}$ controls the emphasis on difficult candidates. For valid spectrum groups $\mathcal G_i$ and batch size $N_b$, equal weighting yields
\begin{equation}
\mathcal L_{\rm SGR}=\frac{1}{N_b}\sum_{i=1}^{N_b}
\frac{\sum_{m\in\mathcal G_i}\ell_{im}}{\max(1,|\mathcal G_i|)}.
\label{eq:rank}
\end{equation}

Grouping constrains positive--negative rankings within each gallery spectrum. Averaging positives and then weighting valid groups equally avoids weighting a spectrum more heavily merely because it has more candidates. SGR acts directly on the inference descriptor without requiring identical feature distributions across spectra. 

\subsection{Optimization Objective}
\label{sec:optimization}
We jointly optimize identity learning, selective local supervision, and cross-scene descriptor ranking.

We combine identity classification, triplet learning, and image--text supervision~\cite{li2022clipreid} in $\mathcal L_{\rm identity}$. The LIA objective $\mathcal L_{\rm LIA}$ combines local classification, selective correspondence alignment, attention diversity, and local-to-global distillation. Classification preserves identity cues, while the alignment and distillation terms use historical observations for complementary forms of supervision.

\paragraph{Joint objective.}
The complete Stage 2 objective is
\begin{equation}
\mathcal L_{\rm total}=\mathcal L_{\rm identity}
+\mathcal L_{\rm LIA}+\lambda_{\rm SGR}(e)\mathcal L_{\rm SGR}.
\label{eq:total}
\end{equation}
OCS constructs the training batches, and $e$ denotes the training epoch. The joint objective connects historical local evidence with current cross-camera rankings through the shared visual representation.

\begin{table*}[!t]
\caption{Comparison under the AS-ReID setting on WHU-MARS. Models are trained and evaluated
separately on WHU-MARS-1000 and WHU-MARS-2337. In WHU-MARS-1000-GD, the WHU-
MARS-1000-trained model is evaluated only on ground-view daytime queries and galleries.}
\label{tab:official}\centering\small
\setlength{\tabcolsep}{1.5pt}
\setlength{\aboverulesep}{0pt}
\setlength{\belowrulesep}{0pt}
\renewcommand{\arraystretch}{1.10}
\begin{tabularx}{\textwidth}{@{}lc*{12}{>{\centering\arraybackslash}X}@{}}\toprule
\rowcolor{tableheader}
& & \multicolumn{4}{c}{\textbf{WHU-MARS-1000-GD}} & \multicolumn{4}{c}{\textbf{WHU-MARS-1000}} & \multicolumn{4}{c}{\textbf{WHU-MARS-2337}}\\
\rowcolor{tableheader}
\multirow{-2}{*}{\textbf{Method}} & \multirow{-2}{*}{\textbf{Venue}} & \textbf{mAP} & \textbf{R-1} & \textbf{R-5} & \textbf{R-10} & \textbf{mAP} & \textbf{R-1} & \textbf{R-5} & \textbf{R-10} & \textbf{mAP} & \textbf{R-1} & \textbf{R-5} & \textbf{R-10}\\\midrule
BoT~\cite{luo2019bot} & CVPRW19 & 10.60 & 24.90 & 37.90 & 45.60 & 5.90 & 18.40 & 32.50 & 40.10 & 4.60 & 14.80 & 26.20 & 32.50\\
TransReID~\cite{he2021transreid} & ICCV21 & 9.60 & 18.90 & 34.10 & 43.40 & 7.40 & 17.10 & 30.30 & 38.20 & 5.50 & 13.10 & 23.40 & 29.10\\
TransReID-SSL~\cite{luo2021transreidssl} & arXiv21 & 14.00 & 30.10 & 44.10 & 51.70 & 10.10 & 26.70 & 43.80 & 51.70 & 9.20 & 24.40 & 39.10 & 46.20\\
AGW~\cite{ye2022survey} & TPAMI22 & 6.10 & 17.00 & 27.80 & 35.90 & 4.00 & 15.60 & 28.70 & 36.00 & 5.20 & 16.90 & 28.70 & 35.10\\
CLIP-ReID~\cite{li2022clipreid} & AAAI23 & 13.80 & 28.90 & 42.70 & 49.60 & 10.60 & 26.60 & 42.60 & 50.40 & 9.30 & 24.10 & 37.50 & 44.00\\
DC-Former~\cite{li2023dcformer} & AAAI23 & 9.90 & 20.10 & 34.90 & 42.00 & 6.50 & 16.30 & 29.90 & 36.70 & 4.30 & 10.90 & 19.80 & 25.50\\
PHA~\cite{zhang2023pha} & CVPR23 & 9.60 & 22.00 & 33.90 & 42.30 & 7.50 & 15.90 & 28.20 & 35.30 & 5.50 & 12.00 & 21.10 & 26.30\\
MSINet~\cite{gu2023msinet} & CVPR23 & 9.70 & 26.90 & 39.50 & 46.30 & 5.60 & 21.10 & 37.50 & 46.00 & 2.60 & 14.80 & 25.40 & 31.60\\
VDT~\cite{zhang2024vdt} & CVPR24 & 13.10 & 31.00 & 44.10 & 51.10 & 7.80 & 25.20 & 41.90 & 50.40 & 6.30 & 20.00 & 33.50 & 40.40\\
SeCap~\cite{wang2025secap} & CVPR25 & 13.40 & 28.10 & 41.20 & 47.80 & 10.40 & 26.40 & 42.40 & 50.40 & 8.00 & 21.40 & 34.10 & 41.00\\
DTST~\cite{wang2024dynamic} & ICME25 & 11.60 & 29.80 & 42.50 & 47.90 & 6.80 & 23.40 & 39.80 & 47.90 & 7.50 & 23.50 & 37.90 & 45.30\\
UAD~\cite{zhao2026whumars} & CVPR26 & 14.50 & 31.90 & 45.00 & 52.50 & 11.00 & 29.50 & 46.30 & 54.80 & 9.60 & 25.70 & 40.90 & 48.20\\
\midrule
\rowcolor{tablehighlight}
OSLA & -- & \textbf{16.57} & \textbf{33.23} & \textbf{47.76} & \textbf{55.28} & \textbf{14.60} & \textbf{34.47} & \textbf{51.07} & \textbf{59.33} & \textbf{12.87} & \textbf{30.49} & \textbf{45.29} & \textbf{52.40}\\
\bottomrule
\end{tabularx}
\end{table*}

\section{Experiments}
\afterpage{\afterpage{\begin{table}[t]
\caption{Comparison under the AG-ReID setting.}
\label{tab:platform}\centering\footnotesize
\setlength{\tabcolsep}{1.5pt}
\setlength{\aboverulesep}{0pt}
\setlength{\belowrulesep}{0pt}
\renewcommand{\arraystretch}{1.10}
\nocite{zhang2024vdt,wang2024dynamic,wang2025secap,zhao2026whumars,li2022clipreid}
\begin{tabularx}{\columnwidth}{@{}l*{4}{>{\centering\arraybackslash}X}@{}}\toprule
\rowcolor{tableheader}
& \multicolumn{2}{c}{A$\to$G} & \multicolumn{2}{c}{G$\to$A}\\
\rowcolor{tableheader}
Method & mAP & R-1 & mAP & R-1\\\midrule
CLIP-ReID (\hyperlink{cite.li2022clipreid}{Li et al., 2023}) & 9.86 & 24.59 & 11.19 & 22.88\\
VDT (\hyperlink{cite.zhang2024vdt}{Zhang et al., 2024}) & 8.90 & 30.40 & 10.70 & 30.20\\
DTST (\hyperlink{cite.wang2024dynamic}{Wang \& Pishgar, 2025}) & 7.90 & 30.50 & 9.80 & 28.50\\
SeCap (\hyperlink{cite.wang2025secap}{Wang et al., 2025}) & 10.10 & 26.40 & 12.00 & 25.30\\
UAD (\hyperlink{cite.zhao2026whumars}{Zhao et al., 2026}) & 11.50 & 31.90 & 13.30 & 30.50\\
\rowcolor{tablehighlight}
OSLA & \textbf{13.82} & \textbf{33.41} & \textbf{15.66} & \textbf{31.88}\\
\bottomrule
\end{tabularx}
\end{table}

\begin{table}[t]
\caption{Comparison under the VI-ReID setting.}
\label{tab:spectra}\centering\footnotesize
\setlength{\tabcolsep}{1.5pt}
\setlength{\aboverulesep}{0pt}
\setlength{\belowrulesep}{0pt}
\renewcommand{\arraystretch}{1.10}
\nocite{ye2022survey,ye2021caj,ye2024cajplus,fang2023saai,yu2025csdn,zhang2023deen,lu2023pmt,zhao2026whumars,li2022clipreid}
\begin{tabularx}{\columnwidth}{@{}l*{4}{>{\centering\arraybackslash}X}@{}}\toprule
\rowcolor{tableheader}
& \multicolumn{2}{c}{NIR$\to$RGB} & \multicolumn{2}{c}{TIR$\to$RGB}\\
\rowcolor{tableheader}
Method & R-1 & mAP & R-1 & mAP\\\midrule
CAJ (\hyperlink{cite.ye2021caj}{Ye et al., 2021}) & 25.90 & 16.11 & 7.92 & 4.79\\
AGW (\hyperlink{cite.ye2022survey}{Ye et al., 2022}) & 17.42 & 10.87 & 5.67 & 3.50\\
SAAI (\hyperlink{cite.fang2023saai}{Fang et al., 2023}) & 23.47 & 13.92 & 5.81 & 3.57\\
CLIP-ReID (\hyperlink{cite.li2022clipreid}{Li et al., 2023}) & 25.85 & 16.18 & 9.13 & 6.72\\
DEEN (\hyperlink{cite.zhang2023deen}{Zhang \& Wang, 2023}) & 27.63 & 17.05 & 8.01 & 5.69\\
PMT (\hyperlink{cite.lu2023pmt}{Lu et al., 2023}) & 27.12 & 18.16 & 8.29 & 5.85\\
CAJ+ (\hyperlink{cite.ye2024cajplus}{Ye et al., 2024}) & 25.57 & 14.62 & 7.78 & 4.37\\
CSDN (\hyperlink{cite.yu2025csdn}{Yu et al., 2025}) & 25.11 & 14.61 & 6.89 & 3.84\\
UAD (\hyperlink{cite.zhao2026whumars}{Zhao et al., 2026}) & 25.53 & 16.58 & 9.04 & 6.32\\
\rowcolor{tablehighlight}
OSLA & \textbf{31.76} & \textbf{21.86} & \textbf{10.40} & \textbf{8.07}\\
\bottomrule
\end{tabularx}
\end{table}

\begin{table}[t]
\caption{Modality-pair evaluation of OSLA and
UAD, each using a single model. Rows and
columns indicate the query and gallery modalities.}
\label{tab:modality-matrix}\centering\footnotesize
\setlength{\tabcolsep}{1.5pt}
\setlength{\aboverulesep}{0pt}
\setlength{\belowrulesep}{0pt}
\renewcommand{\arraystretch}{1.08}
\begin{tabularx}{\columnwidth}{@{}lc*{6}{>{\centering\arraybackslash}X}@{}}\toprule
\rowcolor{tableheader}
\multicolumn{2}{c}{G} & \multicolumn{2}{c}{RGB} & \multicolumn{2}{c}{NIR} & \multicolumn{2}{c}{TIR}\\
\rowcolor{tableheader}
Q & & OSLA & UAD & OSLA & UAD & OSLA & UAD\\\midrule
\multirow{3}{*}{RGB} & R-1 & \textbf{46.37} & 40.47 & \textbf{34.00} & 26.56 & \textbf{11.24} & 8.10\\
& mAP & \textbf{32.09} & 26.83 & \textbf{21.34} & 16.52 & \textbf{7.55} & 6.00\\
& mINP & \textbf{7.33} & 5.36 & \textbf{3.66} & 2.43 & \textbf{1.63} & 1.11\\
\midrule
\multirow{3}{*}{NIR} & R-1 & \textbf{31.76} & 25.53 & \textbf{35.13} & 29.79 & \textbf{10.96} & 7.78\\
& mAP & \textbf{21.86} & 16.58 & \textbf{22.69} & 17.93 & \textbf{7.80} & 5.87\\
& mINP & \textbf{4.36} & 3.10 & \textbf{3.69} & 2.64 & \textbf{1.72} & 1.11\\
\midrule
\multirow{3}{*}{TIR} & R-1 & \textbf{10.40} & 9.04 & \textbf{10.26} & 8.57 & \textbf{22.67} & 20.98\\
& mAP & \textbf{8.07} & 6.32 & \textbf{7.71} & 5.69 & \textbf{15.23} & 14.00\\
& mINP & \textbf{1.85} & 1.36 & \textbf{1.60} & 1.02 & \textbf{3.04} & 2.35\\
\bottomrule
\end{tabularx}
\end{table}

}}
\label{sec:experiments}
\subsection{Experimental Settings}
\label{sec:settings}
\paragraph{Dataset and evaluation protocol.}
WHU-MARS-1000~\cite{zhao2026whumars} has 500 training and 500 test identities, with 6,405 queries and 93,609 gallery images. WHU-MARS-2337 has 1,000 training and 1,337 test identities, with 15,020 queries and 217,116 gallery images. All AS queries are valid. We report mAP, Rank-1/5/10, and mean inverse negative penalty (mINP)~\cite{ye2022survey} in percent, using cosine similarity of normalized 1,280-D descriptors without reranking.

\subsection{Experimental Analysis}
\label{sec:comparison}
We evaluate OSLA under AS-ReID, AG-ReID, and VI-ReID to examine mixed-gallery retrieval, cross-view matching, and cross-spectrum matching, respectively. All settings use the same jointly trained model for each WHU-MARS split. We further evaluate all nine query--gallery spectrum pairs and compare the LIA and CSR configurations to examine their individual and combined performance.

\paragraph{AS-ReID.}
As shown in Table~\ref{tab:official}, OSLA achieves the highest listed mAP and Rank-1/5/10 on both WHU-MARS splits. On WHU-MARS-1000, it obtains 14.60\% mAP and 34.47\% Rank-1, exceeding reported UAD by 3.60 and 4.97 percentage points, respectively. On WHU-MARS-2337, each model is trained and evaluated on the larger split; OSLA reaches 12.87\% mAP and 30.49\% Rank-1, with corresponding gains of 3.27 and 4.79 points. The reported CLIP-ReID results on WHU-MARS-1000 are 10.60\% mAP and 26.60\% Rank-1.

The gains extend beyond the first retrieved image. On WHU-MARS-1000, OSLA reaches 51.07\% Rank-5 and 59.33\% Rank-10, compared with the reported UAD values of 46.30\% and 54.80\%. On the larger 2337 split, the corresponding scores are 45.29\% and 52.40\%.

\paragraph{Ground-daytime subset.}
The GD setting evaluates the WHU-MARS-1000-trained model on ground-daytime queries and galleries. As shown in Table~\ref{tab:official}, OSLA achieves 16.57\% mAP and 33.23\% Rank-1, compared with 14.50\% and 31.90\% for UAD. The gains are 2.07 and 1.33 percentage points, respectively. OSLA also obtains 47.76\% Rank-5 and 55.28\% Rank-10, indicating improved retrieval across multiple ranking positions.

\paragraph{AG-ReID.}
\label{sec:retrievalsettings}
\label{sec:ag}
We evaluate cross-view retrieval in both aerial-to-ground (A$\to$G) and ground-to-aerial (G$\to$A) directions, retaining all spectra in the query and gallery sets (Table~\ref{tab:platform}). OSLA obtains 13.82\% mAP and 33.41\% Rank-1 for A$\to$G, exceeding reported UAD by 2.32 and 1.51 points. For G$\to$A, it reaches 15.66\% mAP and 31.88\% Rank-1, with corresponding gains of 2.36 and 1.38 points. OSLA achieves the highest listed mAP and Rank-1 in both directions. The gains extend to aerial-gallery retrieval, indicating that the learned representation supports cross-view matching without choosing a fixed platform as the retrieval target.

\paragraph{VI-ReID.}
\label{sec:vi}
Table~\ref{tab:spectra} compares NIR$\to$RGB and TIR$\to$RGB retrieval. OSLA reuses its joint AS model without training a separate model for either pair. For NIR$\to$RGB, it achieves 31.76\% Rank-1 and 21.86\% mAP, exceeding the strongest listed competitors in the respective metrics, DEEN~\cite{zhang2023deen} and PMT~\cite{lu2023pmt}, by 4.13 and 3.70 points. For TIR$\to$RGB, it obtains 10.40\% Rank-1 and 8.07\% mAP, compared with 9.04\% and 6.32\% for reported UAD.

DEEN and PMT use separate modality-pair models, whereas UAD and OSLA train jointly. The jointly trained OSLA model achieves the highest listed Rank-1 and mAP in both directions. The lower scores for TIR$\to$RGB relative to NIR$\to$RGB highlight the difficulty of thermal-to-visible matching.

\paragraph{Modality-pair analysis.}
\label{sec:modalitypairs}
Table~\ref{tab:modality-matrix} evaluates all nine query--gallery spectrum pairs with one joint model per method. OSLA exceeds reported UAD in Rank-1, mAP, and mINP for every pair. RGB$\to$RGB reaches 46.37\% Rank-1 and 32.09\% mAP. However, cross-spectrum retrieval involving TIR is substantially harder: RGB$\to$NIR reaches 34.00\% Rank-1, whereas RGB$\to$TIR reaches only 11.24\%. Within-spectrum Rank-1 also varies substantially: 46.37\% for RGB, 35.13\% for NIR, and 22.67\% for TIR. Thus, the aggregate gains coexist with a persistent thermal retrieval challenge.

\subsection{Ablation Studies}
\label{sec:ablation}
\begin{table}[!htbp]
\caption{Ablation study of the OSLA framework.}
\label{tab:ablation}\centering\footnotesize
\setlength{\tabcolsep}{3pt}
\setlength{\aboverulesep}{0pt}
\setlength{\belowrulesep}{0pt}
\renewcommand{\arraystretch}{1.10}
\begin{tabularx}{\columnwidth}{@{}l*{6}{>{\centering\arraybackslash}X}@{}}\toprule
\rowcolor{tableheader}
Method & LIA & CSR & mAP & R-1 & R-5 & R-10\\\midrule
Baseline & $\times$ & $\times$ & 10.66 & 26.40 & 42.14 & 49.93\\
LIA variant & $\checkmark$ & $\times$ & 13.87 & 33.36 & 50.32 & 58.31\\
CSR variant & $\times$ & $\checkmark$ & 14.37 & 34.10 & 50.65 & 58.47\\
\rowcolor{tablehighlight}
OSLA & $\checkmark$ & $\checkmark$ & \textbf{14.60} & \textbf{34.47} & \textbf{51.07} & \textbf{59.33}\\
\bottomrule
\end{tabularx}
\end{table}

\paragraph{Component analysis.}
Table~\ref{tab:ablation} examines the LIA and CSR configurations and their combination. LIA selects local correspondences and transfers qualified identity predictions to global features. CSR organizes and optimizes cross-camera ranking relations. All configurations retrieve images using the global descriptor, allowing the comparison to examine training supervision without introducing local matching at inference.

\paragraph{Individual and joint gains.}
The LIA variant achieves 13.87\% mAP and 33.36\% Rank-1, while the CSR variant reaches 14.37\% and 34.10\%, respectively. CSR also obtains higher Rank-5 and Rank-10 than LIA, reaching 50.65\% and 58.47\%. CSR performs better than LIA across all four metrics, showing the importance of supplying and optimizing cross-camera relations in this setting.

Combining LIA and CSR yields 14.60\% mAP and 34.47\% Rank-1. Relative to CSR, OSLA improves mAP, Rank-1, Rank-5, and Rank-10 by 0.23, 0.37, 0.42, and 0.86 percentage points, respectively. Relative to LIA, the mAP and Rank-1 gains are 0.73 and 1.11 points. OSLA achieves the highest values in all four columns, supporting the complementary use of selective local supervision and cross-scene relation learning. The baseline row provides a reference model trained independently; it is trained without any additional modules.

\subsection{Visualization Analysis}
\label{sec:visualization}
We examine identity separation and cross-scene matching through distance distributions, t-SNE embeddings, and selected retrieval examples.

\begin{figure*}[!tbp]
\centering
\includegraphics[width=\textwidth]{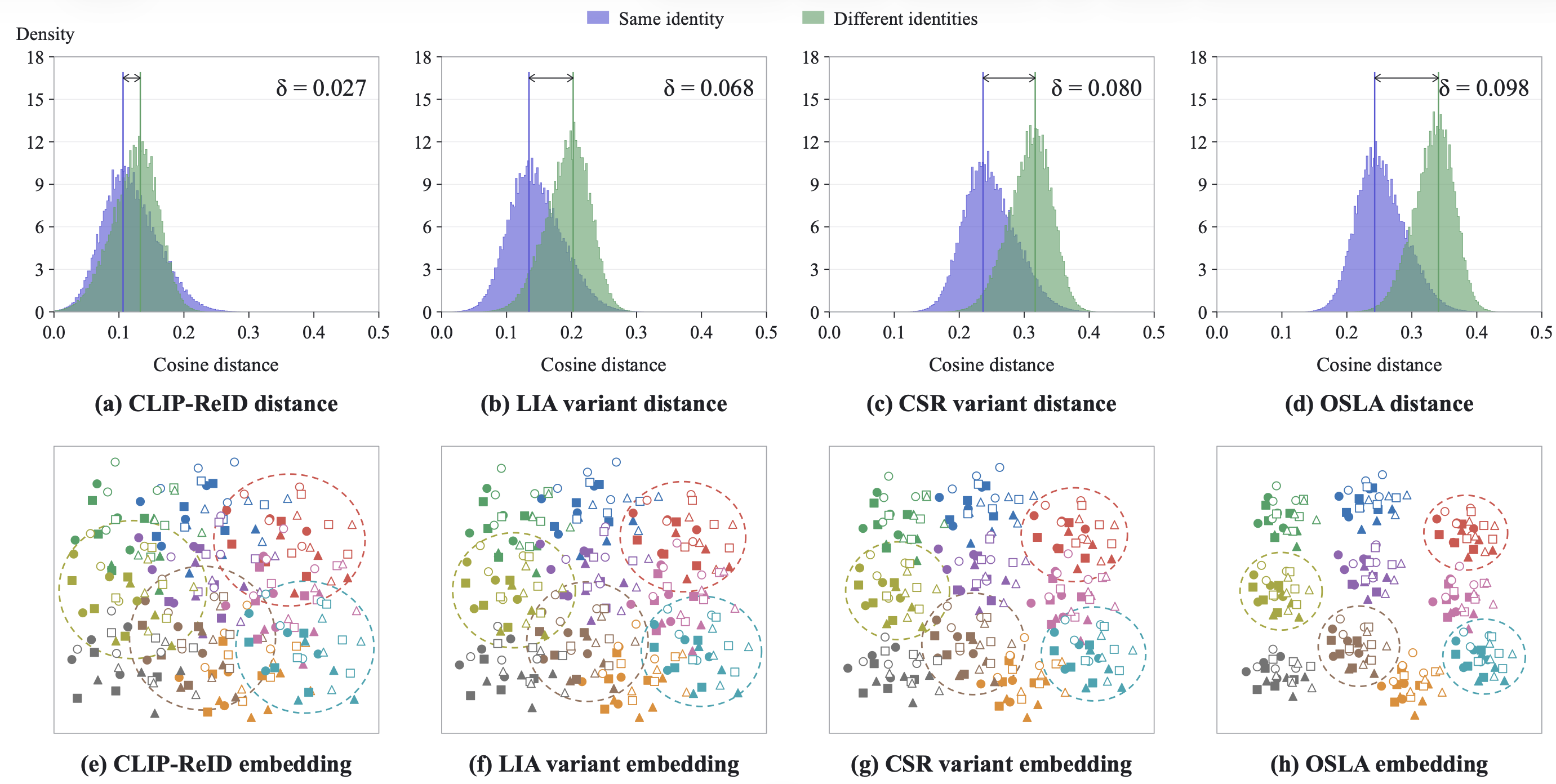}
\caption{Feature analysis: CLIP-ReID, LIA, CSR and OSLA. Top: same-identity (purple) and different-identity (green) cross-camera, cross-spectrum distance densities; vertical lines mark means and $\delta$ their difference. Bottom: colors denote identities; circles/squares/triangles denote RGB/NIR/TIR; filled/open markers denote ground/aerial views. t-SNE panels are fitted independently. Dashed outlines highlight four identities selected for compactness in OSLA, not representative evidence.}
\label{fig:feature-analysis}
\end{figure*}
\begin{figure*}[!tbp]
\centering
\includegraphics[width=\textwidth]{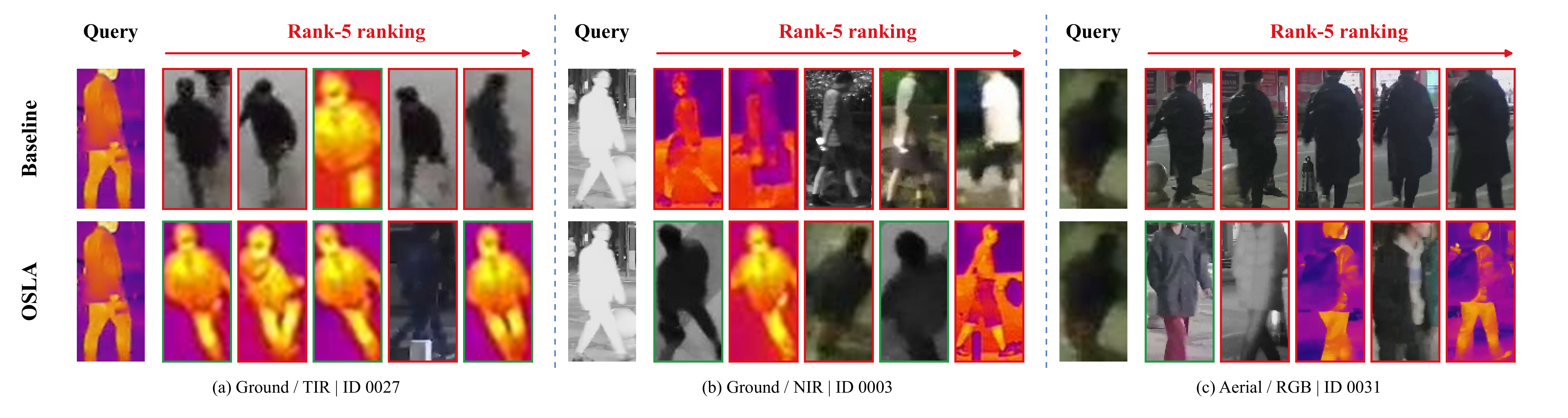}
\caption{Top-5 retrieval for (a) ground TIR, (b) ground NIR and (c) aerial RGB. Each panel compares the shared-recipe baseline (top) with OSLA (bottom). Green/red borders indicate correct/incorrect identities.}
\label{fig:retrieval-main}
\end{figure*}

\paragraph{Feature distributions.}
The top row of Fig.~\ref{fig:feature-analysis} shows cross-camera, cross-spectrum distance distributions. The mean inter- minus intra-identity distance gaps are 0.027, 0.068, 0.080, and 0.098 for CLIP-ReID, LIA, CSR, and OSLA, respectively. OSLA increases the mean distance gap from 0.080 for CSR to 0.098. The larger gap shows increased separation between the two means, rather than uniformly smaller same-identity distances. The distributions still overlap, indicating that difficult positive and negative pairs remain.

In the bottom row, same-color points appear more concentrated and different-color groups are more clearly separated in the OSLA panel. The four dashed outlines highlight selected identity groups that appear more compact under OSLA. These selected groups illustrate the identity structure qualitatively; absolute coordinates are not comparable across the independently fitted panels.

\paragraph{Retrieval examples.}
Figure~\ref{fig:retrieval-main} compares the shared-recipe baseline and OSLA for three selected ground-TIR, ground-NIR, and aerial-RGB queries. OSLA places a correct match first in all three examples. For the ground-TIR query, the first correct match moves from Rank-3 to Rank-1. The number of correct matches in its top-five list also increases from one to four. For the ground-NIR and aerial-RGB queries, the baseline retrieves no correct match within the top five, whereas OSLA retrieves correct matches at Ranks 1, 3, and 4, and at Ranks 1 and 2, respectively. All three queries use the same descriptor without modality-specific model selection. These selected cases illustrate improvements across different acquisition conditions, although incorrect candidates remain in the lists.

The retrieval examples also distinguish two effects of training. For ground TIR, a positive already appears in the baseline list, and OSLA improves both its position and the number of correct results. For ground NIR and aerial RGB, OSLA brings positives into lists that previously contained none. The aerial query retrieves correct ground images in different spectra, illustrating simultaneous platform and spectrum changes. These examples illustrate cross-scene ranking improvements without isolating either module.

\paragraph{Retrieval across acquisition conditions.}
The modality matrix provides a broader context for these examples. NIR$\to$RGB and RGB$\to$NIR reach 31.76\% and 34.00\% Rank-1, while their mAP values are 21.86\% and 21.34\%. Their average precision is comparable despite different Rank-1. By contrast, RGB$\to$TIR and TIR$\to$RGB remain at 11.24\% and 10.40\% Rank-1. The thermal discrepancy is not confined to one choice of query spectrum. 

The component comparison further supports combining the two forms of supervision. CSR exceeds LIA across all four metrics, and OSLA improves each metric further. Relative to CSR, the Rank-10 gain of 0.86 points accompanies the Rank-1 gain of 0.37 points, indicating benefits beyond the first candidate. These comparisons assess the complete modules rather than individual matching rules or OCS and SGR separately.

\section{Conclusion}
\label{sec:conclusion}
In this paper, we propose OSLA for AS-ReID by addressing the reliability of local supervision and the availability of cross-camera training relations. LIA selects supported historical local correspondences, permits unmatched regions, and independently distills qualified local identity predictions into global features. CSR couples observed-cell sampling with spectrum-grouped ranking to supply and optimize same- and cross-spectrum relations on the final descriptor. Both modules are training-only, using a single visual encoder and global retrieval descriptor at inference.

Experiments yield 14.60\% mAP and 34.47\% Rank-1 on WHU-MARS-1000, and 12.87\% mAP and 30.49\% Rank-1 on WHU-MARS-2337. Component comparisons support combining local supervision with cross-scene relation learning. These results extend across retrieval settings with one descriptor. Future work will address incomplete observation coverage and difficult thermal cross-spectrum retrieval.

\section*{LLM Usage Statement}
OpenAI Codex assisted with literature collection, auxiliary code development. The authors designed the study, conducted the experiments, verified the data and revisions, and take responsibility for all scientific claims and interpretations.

{\bibliographystyle{ieeenat_fullname}\bibliography{main}}

\begin{thebibliography}{48}
\providecommand{\natexlab}[1]{#1}
\providecommand{\url}[1]{\texttt{#1}}
\expandafter\ifx\csname urlstyle\endcsname\relax
  \providecommand{\doi}[1]{doi: #1}\else
  \providecommand{\doi}{doi: \begingroup \urlstyle{rm}\Url}\fi

\bibitem[Cuturi(2013)]{cuturi2013sinkhorn}
Marco Cuturi.
\newblock Sinkhorn distances: Lightspeed computation of optimal transportation
  distances.
\newblock In \emph{Advances in Neural Information Processing Systems}, pages
  2292--2300, 2013.

\bibitem[Dosovitskiy et~al.(2021)Dosovitskiy, Beyer, Kolesnikov, Weissenborn,
  Zhai, Unterthiner, Dehghani, Minderer, Heigold, Gelly, Uszkoreit, and
  Houlsby]{dosovitskiy2021vit}
Alexey Dosovitskiy, Lucas Beyer, Alexander Kolesnikov, Dirk Weissenborn,
  Xiaohua Zhai, Thomas Unterthiner, Mostafa Dehghani, Matthias Minderer, Georg
  Heigold, Sylvain Gelly, Jakob Uszkoreit, and Neil Houlsby.
\newblock An image is worth 16x16 words: Transformers for image recognition at
  scale.
\newblock In \emph{ICLR}, 2021.

\bibitem[Fang et~al.(2023)Fang, Yang, and Fu]{fang2023saai}
Xingye Fang, Yang Yang, and Ying Fu.
\newblock Visible-infrared person re-identification via semantic alignment and
  affinity inference.
\newblock In \emph{ICCV}, pages 11270--11279, 2023.

\bibitem[Gu et~al.(2023)Gu, Wang, Luo, Chen, Jiang, Fang, Zhang, You, and
  Zhao]{gu2023msinet}
Jianyang Gu, Kai Wang, Hao Luo, Chen Chen, Wei Jiang, Yuqiang Fang, Shanghang
  Zhang, Yang You, and Jian Zhao.
\newblock {MSINet}: Twins contrastive search of multi-scale interaction for
  object {ReID}.
\newblock In \emph{CVPR}, pages 19243--19253, 2023.

\bibitem[Ha et~al.(2025)Ha, Jiang, Li, Pan, Zhu, Zhang, Zhu, Gong, and
  Wang]{ha2025multi}
Ruiyang Ha, Songyi Jiang, Bin Li, Bikang Pan, Yihang Zhu, Junjie Zhang, Xiatian
  Zhu, Shaogang Gong, and Jingya Wang.
\newblock Multi-modal multi-platform person re-identification: Benchmark and
  method.
\newblock \emph{arXiv preprint arXiv:2503.17096}, 2025.

\bibitem[He et~al.(2021)He, Luo, Wang, Wang, Li, and Jiang]{he2021transreid}
Shuting He, Hao Luo, Pichao Wang, Fan Wang, Hao Li, and Wei Jiang.
\newblock {TransReID}: Transformer-based object re-identification.
\newblock In \emph{ICCV}, pages 15013--15022, 2021.

\bibitem[Hermans et~al.(2017)Hermans, Beyer, and Leibe]{hermans2017triplet}
Alexander Hermans, Lucas Beyer, and Bastian Leibe.
\newblock In defense of the triplet loss for person re-identification.
\newblock \emph{arXiv preprint arXiv:1703.07737}, 2017.

\bibitem[Hinton et~al.(2015)Hinton, Vinyals, and Dean]{hinton2015distilling}
Geoffrey Hinton, Oriol Vinyals, and Jeff Dean.
\newblock Distilling the knowledge in a neural network.
\newblock \emph{arXiv preprint arXiv:1503.02531}, 2015.

\bibitem[Jia et~al.(2022)Jia, Tang, Chen, Cardie, Belongie, Hariharan, and
  Lim]{jia2022vpt}
Menglin Jia, Luming Tang, Bor-Chun Chen, Claire Cardie, Serge Belongie, Bharath
  Hariharan, and Ser-Nam Lim.
\newblock Visual prompt tuning.
\newblock In \emph{ECCV}, pages 709--727, 2022.

\bibitem[Khosla et~al.(2020)Khosla, Teterwak, Wang, Sarna, Tian, Isola,
  Maschinot, Liu, and Krishnan]{khosla2020supcon}
Prannay Khosla, Piotr Teterwak, Chen Wang, Aaron Sarna, Yonglong Tian, Phillip
  Isola, Aaron Maschinot, Ce Liu, and Dilip Krishnan.
\newblock Supervised contrastive learning.
\newblock In \emph{Advances in Neural Information Processing Systems}, 2020.

\bibitem[Kim et~al.(2023)Kim, Kim, Park, Park, and Sohn]{kim2023partmix}
Minsu Kim, Seungryong Kim, Jungin Park, Seongheon Park, and Kwanghoon Sohn.
\newblock {PartMix}: Regularization strategy to learn part discovery for
  visible-infrared person re-identification.
\newblock In \emph{CVPR}, pages 18621--18632, 2023.

\bibitem[Li et~al.(2023{\natexlab{a}})Li, Sun, and Li]{li2022clipreid}
Siyuan Li, Li Sun, and Qingli Li.
\newblock {CLIP-ReID}: Exploiting vision-language model for image
  re-identification without concrete text labels.
\newblock In \emph{AAAI}, pages 1405--1413, 2023{\natexlab{a}}.

\bibitem[Li et~al.(2018{\natexlab{a}})Li, Zhu, and Gong]{li2018hacnn}
Wei Li, Xiatian Zhu, and Shaogang Gong.
\newblock {Harmonious Attention Network for Person Re-Identification}.
\newblock In \emph{CVPR}, pages 2285--2294, 2018{\natexlab{a}}.

\bibitem[Li et~al.(2023{\natexlab{b}})Li, Zou, Wang, Xu, Zhao, Zheng, Cheng,
  and Chu]{li2023dcformer}
Wen Li, Cheng Zou, Meng Wang, Furong Xu, Jianan Zhao, Ruobing Zheng, Yuan
  Cheng, and Wei Chu.
\newblock {DC-Former}: Diverse and compact transformer for person
  re-identification.
\newblock In \emph{AAAI}, pages 1415--1423, 2023{\natexlab{b}}.

\bibitem[Li et~al.(2018{\natexlab{b}})Li, Grandvalet, and Davoine]{li2018l2sp}
Xuhong Li, Yves Grandvalet, and Franck Davoine.
\newblock Explicit inductive bias for transfer learning with convolutional
  networks.
\newblock In \emph{ICML}, pages 2825--2834, 2018{\natexlab{b}}.

\bibitem[Li et~al.(2021)Li, He, Zhang, Liu, Zhang, and Wu]{li2021pat}
Yulin Li, Jianfeng He, Tianzhu Zhang, Xiang Liu, Yongdong Zhang, and Feng Wu.
\newblock {Diverse Part Discovery: Occluded Person Re-Identification With
  Part-Aware Transformer}.
\newblock In \emph{CVPR}, pages 2898--2907, 2021.

\bibitem[Lu et~al.(2023)Lu, Zou, and Zhang]{lu2023pmt}
Hu Lu, Xuezhang Zou, and Pingping Zhang.
\newblock Learning progressive modality-shared transformers for effective
  visible-infrared person re-identification.
\newblock In \emph{AAAI}, pages 1835--1843, 2023.

\bibitem[Luo et~al.(2019)Luo, Gu, Liao, Lai, and Jiang]{luo2019bot}
Hao Luo, Youzhi Gu, Xingyu Liao, Shenqi Lai, and Wei Jiang.
\newblock Bag of tricks and a strong baseline for deep person
  re-identification.
\newblock In \emph{CVPR Workshops}, pages 1487--1495, 2019.

\bibitem[Luo et~al.(2021)Luo, Wang, Xu, Ding, Zhou, Wang, Li, and
  Jin]{luo2021transreidssl}
Hao Luo, Pichao Wang, Yi Xu, Feng Ding, Yanxin Zhou, Fan Wang, Hao Li, and Rong
  Jin.
\newblock Self-supervised pre-training for transformer-based person
  re-identification.
\newblock \emph{arXiv preprint arXiv:2111.12084}, 2021.

\bibitem[Nguyen et~al.(2017)Nguyen, Hong, Kim, and Park]{nguyen2017regdb}
Dat~Tien Nguyen, Hyung~Gil Hong, Ki~Wan Kim, and Kang~Ryoung Park.
\newblock Person recognition system based on a combination of body images from
  visible light and thermal cameras.
\newblock \emph{Sensors}, 17\penalty0 (3):\penalty0 605, 2017.

\bibitem[Nguyen et~al.(2025)Nguyen, Nguyen, Pemasiri, Jahan, Fookes, and
  Sridharan]{nguyen2025agvpreidvir}
Huy Nguyen, Kien Nguyen, Akila Pemasiri, Akmal Jahan, Clinton Fookes, and
  Sridha Sridharan.
\newblock {AG-VPReID.VIR}: Bridging aerial and ground platforms for video-based
  visible-infrared person re-id.
\newblock In \emph{IJCB}, pages 1--10, 2025.

\bibitem[Park et~al.(2019)Park, Kim, Lu, and Cho]{park2019rkd}
Wonpyo Park, Dongju Kim, Yan Lu, and Minsu Cho.
\newblock {Relational Knowledge Distillation}.
\newblock In \emph{CVPR}, pages 3962--3971, 2019.

\bibitem[Radford et~al.(2021)Radford, Kim, Hallacy, Ramesh, Goh, Agarwal,
  Sastry, Askell, Mishkin, Clark, Krueger, and Sutskever]{radford2021clip}
Alec Radford, Jong~Wook Kim, Chris Hallacy, Aditya Ramesh, Gabriel Goh,
  Sandhini Agarwal, Girish Sastry, Amanda Askell, Pamela Mishkin, Jack Clark,
  Gretchen Krueger, and Ilya Sutskever.
\newblock Learning transferable visual models from natural language
  supervision.
\newblock In \emph{ICML}, pages 8748--8763, 2021.

\bibitem[Ren and Zhang(2024)]{ren2024idkl}
Kaijie Ren and Lei Zhang.
\newblock Implicit discriminative knowledge learning for visible-infrared
  person re-identification.
\newblock In \emph{CVPR}, pages 393--402, 2024.

\bibitem[Sarlin et~al.(2020)Sarlin, DeTone, Malisiewicz, and
  Rabinovich]{sarlin2020superglue}
Paul-Edouard Sarlin, Daniel DeTone, Tomasz Malisiewicz, and Andrew Rabinovich.
\newblock {SuperGlue}: Learning feature matching with graph neural networks.
\newblock In \emph{CVPR}, pages 4938--4947, 2020.

\bibitem[Schroff et~al.(2015)Schroff, Kalenichenko, and
  Philbin]{schroff2015facenet}
Florian Schroff, Dmitry Kalenichenko, and James Philbin.
\newblock {FaceNet: A Unified Embedding for Face Recognition and Clustering}.
\newblock In \emph{CVPR}, pages 815--823, 2015.

\bibitem[Sun et~al.(2018)Sun, Zheng, Yang, Tian, and Wang]{sun2018pcb}
Yifan Sun, Liang Zheng, Yi Yang, Qi Tian, and Shengjin Wang.
\newblock {Beyond Part Models: Person Retrieval with Refined Part Pooling (and
  A Strong Convolutional Baseline)}.
\newblock In \emph{ECCV}, pages 501--518, 2018.

\bibitem[Sun et~al.(2020)Sun, Cheng, Zhang, Zhang, Zheng, Wang, and
  Wei]{sun2020circle}
Yifan Sun, Changmao Cheng, Yuhan Zhang, Chi Zhang, Liang Zheng, Zhongdao Wang,
  and Yichen Wei.
\newblock Circle loss: A unified perspective of pair similarity optimization.
\newblock In \emph{CVPR}, pages 6398--6407, 2020.

\bibitem[Wang et~al.(2018)Wang, Yuan, Chen, Li, and Zhou]{wang2018mgn}
Guanshuo Wang, Yufeng Yuan, Xiong Chen, Jiwei Li, and Xi Zhou.
\newblock Learning discriminative features with multiple granularities for
  person re-identification.
\newblock In \emph{Proceedings of the 26th ACM International Conference on
  Multimedia}, pages 274--282, 2018.

\bibitem[Wang et~al.(2020{\natexlab{a}})Wang, Yang, Liu, Wang, Yang, Wang, Yu,
  Zhou, and Sun]{wang2020horeid}
Guan'an Wang, Shuo Yang, Huanyu Liu, Zhicheng Wang, Yang Yang, Shuliang Wang,
  Gang Yu, Erjin Zhou, and Jian Sun.
\newblock {High-Order Information Matters: Learning Relation and Topology for
  Occluded Person Re-Identification}.
\newblock In \emph{CVPR}, pages 6448--6457, 2020{\natexlab{a}}.

\bibitem[Wang et~al.(2025)Wang, Wang, Wu, Jiao, Wang, and Wang]{wang2025secap}
Shining Wang, Yunlong Wang, Ruiqi Wu, Bingliang Jiao, Wenxuan Wang, and Peng
  Wang.
\newblock {SeCap}: Self-calibrating and adaptive prompts for cross-view person
  re-identification in aerial-ground networks.
\newblock In \emph{CVPR}, pages 22119--22128, 2025.

\bibitem[Wang et~al.(2019)Wang, Han, Huang, Dong, and
  Scott]{wang2019multisimilarity}
Xun Wang, Xintong Han, Weilin Huang, Dengke Dong, and Matthew~R. Scott.
\newblock Multi-similarity loss with general pair weighting for deep metric
  learning.
\newblock In \emph{CVPR}, pages 5022--5030, 2019.

\bibitem[Wang et~al.(2020{\natexlab{b}})Wang, Zhang, Huang, and
  Scott]{wang2020xbm}
Xun Wang, Haozhi Zhang, Weilin Huang, and Matthew~R. Scott.
\newblock {Cross-Batch Memory for Embedding Learning}.
\newblock In \emph{CVPR}, pages 6387--6396, 2020{\natexlab{b}}.

\bibitem[Wang and Pishgar(2025)]{wang2024dynamic}
Yuhai Wang and Maryam Pishgar.
\newblock Dynamic token selective transformer for aerial-ground person
  re-identification.
\newblock In \emph{ICME}, pages 1--6, 2025.
\newblock Preprint: arXiv:2412.00433, 2024.

\bibitem[Wu et~al.(2017)Wu, Zheng, Yu, Gong, and Lai]{wu2017rgbir}
Ancong Wu, Wei-Shi Zheng, Hong-Xing Yu, Shaogang Gong, and Jianhuang Lai.
\newblock {RGB-Infrared Cross-Modality Person Re-Identification}.
\newblock In \emph{ICCV}, pages 5390--5399, 2017.

\bibitem[Wu et~al.(2021)Wu, Dai, Chen, Lin, Wu, Huang, Zhong, and
  Ji]{wu2021mpanet}
Qiong Wu, Pingyang Dai, Jie Chen, Chia-Wen Lin, Yongjian Wu, Feiyue Huang,
  Bineng Zhong, and Rongrong Ji.
\newblock {Discover Cross-Modality Nuances for Visible-Infrared Person
  Re-Identification}.
\newblock In \emph{CVPR}, pages 4330--4339, 2021.

\bibitem[Ye et~al.(2020)Ye, Shen, Crandall, Shao, and Luo]{ye2020ddag}
Mang Ye, Jianbing Shen, David~J. Crandall, Ling Shao, and Jiebo Luo.
\newblock Dynamic dual-attentive aggregation learning for visible-infrared
  person re-identification.
\newblock In \emph{ECCV}, pages 229--247, 2020.

\bibitem[Ye et~al.(2021)Ye, Ruan, Du, and Shou]{ye2021caj}
Mang Ye, Weijian Ruan, Bo Du, and Mike~Zheng Shou.
\newblock Channel augmented joint learning for visible-infrared recognition.
\newblock In \emph{ICCV}, pages 13567--13576, 2021.

\bibitem[Ye et~al.(2022)Ye, Shen, Lin, Xiang, Shao, and Hoi]{ye2022survey}
Mang Ye, Jianbing Shen, Gaojie Lin, Tao Xiang, Ling Shao, and Steven C.~H. Hoi.
\newblock Deep learning for person re-identification: A survey and outlook.
\newblock \emph{IEEE Transactions on Pattern Analysis and Machine
  Intelligence}, 44\penalty0 (6):\penalty0 2872--2893, 2022.
\newblock Preprint: arXiv:2001.04193, 2020; early access, 2021.

\bibitem[Ye et~al.(2024)Ye, Wu, Chen, and Du]{ye2024cajplus}
Mang Ye, Zesen Wu, Cuiqun Chen, and Bo Du.
\newblock Channel augmentation for visible-infrared re-identification.
\newblock \emph{IEEE Transactions on Pattern Analysis and Machine
  Intelligence}, 46\penalty0 (4):\penalty0 2299--2315, 2024.

\bibitem[Yu et~al.(2025)Yu, Dong, Zhu, Peng, and Tao]{yu2025csdn}
Xiaoyan Yu, Neng Dong, Liehuang Zhu, Hao Peng, and Dapeng Tao.
\newblock {CLIP}-driven semantic discovery network for visible-infrared person
  re-identification.
\newblock \emph{IEEE Transactions on Multimedia}, 27:\penalty0 4137--4150,
  2025.

\bibitem[Zhang et~al.(2023)Zhang, Zhang, Zhang, Li, and Pu]{zhang2023pha}
Guiwei Zhang, Yongfei Zhang, Tianyu Zhang, Bo Li, and Shiliang Pu.
\newblock {PHA}: Patch-wise high-frequency augmentation for transformer-based
  person re-identification.
\newblock In \emph{CVPR}, pages 14133--14142, 2023.

\bibitem[Zhang et~al.(2024)Zhang, Wang, Patel, Xie, and Lai]{zhang2024vdt}
Quan Zhang, Lei Wang, Vishal~M. Patel, Xiaohua Xie, and Jianhuang Lai.
\newblock View-decoupled transformer for person re-identification under
  aerial-ground camera network.
\newblock In \emph{CVPR}, pages 22000--22009, 2024.

\bibitem[Zhang et~al.(2026)Zhang, Cai, Zhao, Wu, Wu, Chen, and
  Lai]{zhang2026visa}
Quan Zhang, Zeqiang Cai, Peiming Zhao, Jingze Wu, Cailun Wu, Hongbo Chen, and
  Jianhuang Lai.
\newblock View-aware semantic alignment for aerial-ground person
  re-identification.
\newblock In \emph{CVPR}, pages 4383--4392, 2026.

\bibitem[Zhang and Wang(2023)]{zhang2023deen}
Yukang Zhang and Hanzi Wang.
\newblock Diverse embedding expansion network and low-light cross-modality
  benchmark for visible-infrared person re-identification.
\newblock In \emph{CVPR}, pages 2153--2162, 2023.

\bibitem[Zhao et~al.(2026)Zhao, Zhou, Yang, Li, Liang, Chen, Du, and
  Ye]{zhao2026whumars}
Yuxuan Zhao, Zhongao Zhou, Bin Yang, He Li, Jian Liang, Jun Chen, Bo Du, and
  Mang Ye.
\newblock {WHU-MARS}: A multispectral aerial-ground benchmark towards
  any-scenario person re-identification.
\newblock In \emph{CVPR}, pages 25461--25471, 2026.

\bibitem[Zhou et~al.(2022{\natexlab{a}})Zhou, Yang, Loy, and
  Liu]{zhou2022cocoop}
Kaiyang Zhou, Jingkang Yang, Chen~Change Loy, and Ziwei Liu.
\newblock {Conditional Prompt Learning for Vision-Language Models}.
\newblock In \emph{CVPR}, pages 16816--16825, 2022{\natexlab{a}}.

\bibitem[Zhou et~al.(2022{\natexlab{b}})Zhou, Yang, Loy, and Liu]{zhou2022coop}
Kaiyang Zhou, Jingkang Yang, Chen~Change Loy, and Ziwei Liu.
\newblock Learning to prompt for vision-language models.
\newblock \emph{International Journal of Computer Vision}, 130\penalty0
  (9):\penalty0 2337--2348, 2022{\natexlab{b}}.

\end{thebibliography}
\clearpage

\appendix
\setcounter{equation}{0}
\renewcommand{\theequation}{S\arabic{equation}}
\renewcommand{\theHequation}{S\arabic{equation}}
\setcounter{figure}{0}
\renewcommand{\thefigure}{S\arabic{figure}}
\setcounter{table}{0}
\renewcommand{\thetable}{S\arabic{table}}
\allowdisplaybreaks[4]

\section{Implementation Details}
\label{app:implementation}
\paragraph{Training schedule and augmentation.}
We optimize both stages with Adam and batch size 64. Stage 1 runs for 120 epochs with peak learning rate $3.5\times10^{-4}$. Stage 2 runs for 60 epochs with learning rate $5\times10^{-6}$ and weight decay $10^{-4}$, using ten warmup epochs and learning-rate decays by 0.1 at epochs 30 and 50. Augmentation combines horizontal flipping, padded cropping, and random erasing.

\paragraph{Loss weights and reductions.}
We use $\lambda_{\rm loc}=0.1$. Distillation and ranking weights rise from zero after epoch 5 to 0.05 over five and ten epochs, respectively. The LIA objective weights local classification, alignment, and diversity by $1$, $0.5$, and $0.1$ inside the local term. Alignment uses $\max(\sum_{s',j}\rho_{y_is'j},10^{-12})$ as its numerical denominator. Images without eligible references do not enter the alignment reduction. Teachers are stop-gradient; historical memory and correspondence assignments are detached. An SGR anchor without a valid spectrum group contributes zero but remains in the batch denominator.

\paragraph{Candidate selection and sampling.}
Competing identities come from the current batch and must have at least two observations in the relevant memory scene. A correspondence is rejected if no eligible competitor exists. Candidate and unmatched logits are $C_{kj}/\tau_M$ and $\eta/\tau_M$, respectively; disallowed pairs are masked. OCS prioritizes cross-platform camera pairs when available and otherwise uses two ground cameras. One image is drawn from each of the four selected observed cells, and spectrum pairs rotate across identity visits.

\paragraph{Local features and memory.}
We use $K=4$ free queries, $\tau_q=0.07$, and three spectrum-specific bias-free linear classifiers applied to $16\unit u_k$. Their weights are initialized from the 768-D global identity classifier. For scene $s=(p,m)$, queries are $q_{ks}=q_k+d_{km}+e_{kp}$, combining shared vectors with spectrum and platform offsets. Query offsets are initially zero. The attention diversity term is the mean off-diagonal cosine similarity between attention rows, averaged over images; local CE, alignment, and KL use the scene-balanced reductions specified in the main text.

An identity--scene memory provides cross-batch references by storing historical local feature centers $M_{ysk}$, accumulated observation counts $n_{ys}$, and feature dispersion $d_{ysk}$. Centers are normalized. Dispersion reflects both within-update spread and drift relative to the historical center. Current supervision uses historical memory.

For a successful update, let $\mathcal B_{ys}$ contain the images of identity $y$ in scene $s$. For each local slot, compute
\begin{align}
\widehat M_{ysk}&=\operatorname{normalize}\left(\frac{1}{|\mathcal B_{ys}|}
\sum_{i\in\mathcal B_{ys}}\unit u_{ik}\right),\\
\widehat d_{ysk}&=\frac{1}{|\mathcal B_{ys}|}\sum_{i\in\mathcal B_{ys}}
\clip(1-\unit u_{ik}^{\top}\widehat M_{ysk},0,2).
\end{align}
For an existing center, with $\gamma=0.2$ and drift
$\Delta_{ysk}=\clip(1-\widehat M_{ysk}^{\top}M_{ysk},0,2)$, the updates are
\begin{align}
M_{ysk}^{+}&=\operatorname{normalize}((1-\gamma)M_{ysk}+\gamma\widehat M_{ysk}),\\
d_{ysk}^{+}&=(1-\gamma)d_{ysk}+\tfrac{\gamma}{2}(\widehat d_{ysk}+\Delta_{ysk}).
\end{align}
A first observation initializes the center and dispersion directly with $\widehat M$ and $\widehat d$. The count increases by $|\mathcal B_{ys}|$, counting image observations rather than unique source images. Updates skipped by automatic mixed precision do not change the memory. Targets and correspondence decisions always use the previous memory state.

A foreign scene $s'\ne s$ with sufficient observation support provides references with source weight
\begin{equation}
\rho_{ys'j}=\sqrt{\frac{\min(n_{ys'},n_{\max})}{n_{\max}}}
\exp(-\beta d_{ys'j}).
\label{eq:supp-reliability}
\end{equation}
Here $n_{\max}$ caps observation support and $\beta$ controls the influence of dispersion. This weighting favors repeatedly observed, stable references rather than treating every stored feature as equally reliable.

\begin{figure*}[!t]
\centering
\includegraphics[width=0.8\textwidth]{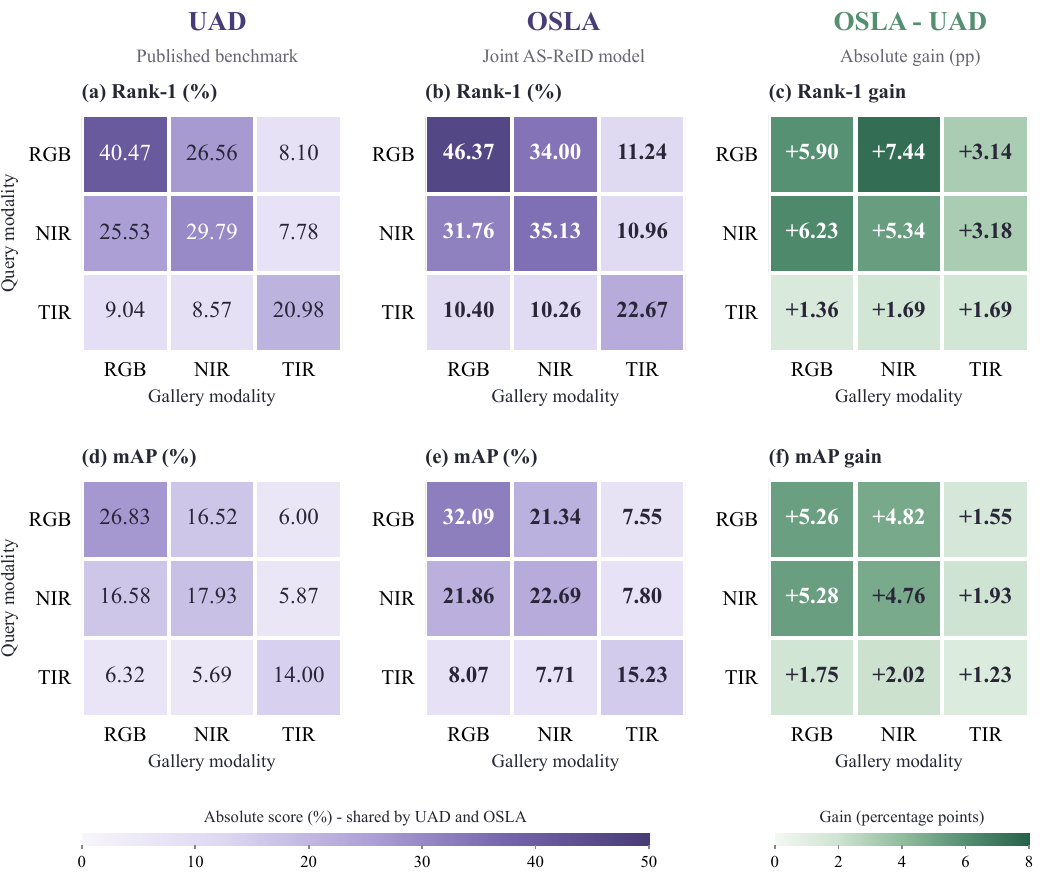}
\setcounter{figure}{1}
\caption{Modality-pair comparison on the official WHU-MARS-1000 split. Top: Rank-1; bottom: mAP. Rows within each matrix indicate query modalities and columns indicate gallery modalities. Purple panels share an absolute-score scale (\%); green panels show OSLA minus UAD in percentage points (pp).}
\label{fig:uad-osla-modality-heatmap}
\end{figure*}

\paragraph{Matching and transfer.}
The source-count threshold is two observations, $n_{\max}=64$, and $\beta=4$. Matching uses $\eta=0.2$, $\epsilon_{\rm id}=0.01$, $\tau_M=0.05$, and 30 log-domain Sinkhorn iterations. After masking disallowed pairs, the real plan is divided first by row sums clamped below at one, then by column sums clamped below at one. This enforces both capacities at finite iteration count. The alignment normalization retains the reliability of rejected source slots. Teachers use $T_{\rm KD}=2$ and require correct prior-center classification; accepted transport mass is not a teacher condition.

\paragraph{Descriptor ranking.}
CSR couples observed-cell sampling (OCS) with spectrum-grouped ranking (SGR). SGR uses $\tau_{\rm SGR}=0.07$ and margin $\mu_{\rm SGR}=0.1$. Positive and negative membership both require different camera IDs from the anchor. Spectrum groups are valid only when both sets are nonempty. The descriptor is formed from raw $g$ and $v$, concatenated and then normalized, exactly as at inference. OCS uses real image indices in the four observed cells and cycles spectrum pairs across identity visits.

\section{Training Objectives}
\paragraph{Identity learning objective.}
We combine identity classification, triplet learning, and image--text supervision:
\begin{equation}
\mathcal L_{\rm identity}=0.25\mathcal L_{\rm ID}
+\mathcal L_{\rm tri}+\mathcal L_{\rm I2T}.
\label{eq:supp-identity-full}
\end{equation}
Here $\mathcal L_{\rm ID}$ provides label-smoothed identity supervision, $\mathcal L_{\rm tri}$ encourages identity separation through batch-hard triplet learning, and $\mathcal L_{\rm I2T}$ relates projected image features to identity text features.

\paragraph{Local objective.}
The LIA objective is
\begin{equation}
\begin{split}
\mathcal L_{\rm LIA}={}&\lambda_{\rm loc}(\mathcal L_{\rm localCE}
+0.5\mathcal L_{\rm align}+0.1\mathcal L_{\rm div})\\
&+\lambda_{\rm KD}(e)\mathcal L_{\rm KL}.
\end{split}
\label{eq:supp-local-full}
\end{equation}
The terms are local identity classification, correspondence alignment, attention-overlap regularization, and knowledge distillation, respectively.

\begin{align}
\mathcal L_{\rm ID}&=\CE_{\rm ls}(a^g,y)+\CE_{\rm ls}(a^v,y),\label{eq:id}\\
\mathcal L_{\rm tri}&=\mathcal T(g^{-},y)+\mathcal T(g,y)+\mathcal T(v,y),\label{eq:tri}\\
\mathcal L_{\rm I2T}&=\CE_{\rm ls}\big((v^\top t_j)_{j=1}^{C},y\big),\label{eq:i2t}
\end{align}
where $a^g$ and $a^v$ are the two BN-neck identity logits, $t_j$ is the frozen Stage 1 identity text feature, and $C$ is the number of training identities. $\CE_{\rm ls}$ uses label smoothing of 0.1, and $\mathcal T$ is batch-hard triplet loss with margin 0.3 on the raw branch features. Branch losses are summed, not averaged; image-to-text classification uses the projected-feature dot product.

\section{Early LIA Training Dynamics}
\label{app:lia-dynamics}

\begin{figure}[!ht]
\centering
\includegraphics[width=\columnwidth]{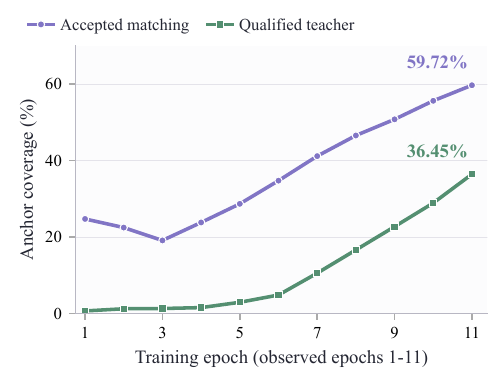}
\setcounter{figure}{0}
\caption{Early LIA dynamics on WHU-MARS-1000 (500 training identities, seed 1). Curves show epoch-level mean matching and teacher coverage over attempted batches, including AMP-skipped attempts. Only observed epochs 1--11 are shown.}
\label{fig:supp-lia-early}
\end{figure}

We inspect the saved diagnostic record of the official OSLA run. Matching coverage is the fraction of batch anchors with accepted local matching; teacher coverage is the fraction with at least one qualified historical local teacher. Figure~\ref{fig:supp-lia-early} averages these fractions over all attempted batches in each observed epoch. Memory updates occur only after successful optimization steps.

Matching coverage changes from 24.76\% in epoch 1 to 19.14\% in epoch 3, then increases to 59.72\% in epoch 11. Teacher coverage increases from 0.70\% to 36.45\%. Historical supervision is therefore selective, with availability changing during training. Matching and teachers are selected independently: a qualified teacher need not have an accepted correspondence. Marginal coverages do not measure their overlap or correspondence accuracy.

The available record covers epochs 1--11 only, not convergence across all 60 epochs. No later values are extrapolated, and these diagnostics do not replace ablations of the selection rules.

\section{Modality-Pair Comparison}
\label{app:modality-heatmap}
Figure~\ref{fig:uad-osla-modality-heatmap} visualizes the modality-pair results reported in the main paper. Evaluation uses the official WHU-MARS-1000 split with 500 training and 500 test identities. Each pair contains 2,135 valid queries and a 31,203-image gallery before same-camera exclusion.

\paragraph{Magnitude and direction of the gains.}
OSLA has higher Rank-1 and mAP than UAD values for all nine pairs. The largest Rank-1 gain is RGB-to-NIR, from 26.56\% to 34.00\% (+7.44 pp); NIR-to-RGB improves from 25.53\% to 31.76\% (+6.23 pp). Their mAP gains are +4.82 and +5.28 pp, respectively. Both directions improve first-match retrieval and the ordering of relevant gallery images. Their gains differ because swapping modalities changes the query and gallery observations.

\paragraph{Remaining cross-spectrum difficulty.}
Retrieval involving TIR remains difficult, particularly across spectra. RGB-to-TIR and NIR-to-TIR gain +3.14 and +3.18 pp in Rank-1, yet their absolute scores remain 11.24\% and 10.96\%. In the reverse directions, TIR-to-RGB and TIR-to-NIR reach 10.40\% and 10.26\%, with mAP of 8.07\% and 7.71\%. Thus, the gain panels should be read alongside the absolute scores: the cross-spectrum gap remains substantial. 

\end{document}